%% file: acl_latex.tex
\pdfoutput=1
\documentclass[11pt]{article}

\usepackage[final]{acl}

\usepackage{times}
\usepackage{latexsym}

\usepackage[T1]{fontenc}
\usepackage[utf8]{inputenc}
\usepackage{float}
\usepackage{microtype}
\input{definition}
\input{math_comm}

\title{On Robustness of Prompt-based Semantic Parsing \\with Large Pre-trained Language Model:\\An Empirical Study on Codex}

\author{\bf{Terry Yue Zhuo}$^{1,2}$ \and \bf{Zhuang Li}$^2$\thanks{~~corresponding author} \\ \bf{Yujin Huang}$^2$ \and \bf{Fatemeh Shiri}$^2$ \\ \bf{Weiqing Wang}$^2$ \and \bf{Gholamreza Haffari}$^2$ \and \bf{Yuan-Fang Li}$^2$ \\
         $^1$CSIRO's Data61, Australia\\
         $^2$Monash University, Australia\\
         \tt \{terry.zhuo, zhuang.li\}@monash.edu
         }

\begin{document}

\maketitle
\input{sections/abstract}
\input{sections/01-introduction}
\input{sections/05-related-work}

\input{sections/02-attack-method}

\input{sections/03-weakly-adversarial-guidence}
\input{sections/04-experiment}
\input{sections/06-conclusion}

\section*{Limitations}
In this study, we examine the robustness of the prompt-based semantic parser, \codex, by focusing on the impact of prompt design on its execution performance. However, there is a need for future research to investigate more alternative adversarial training strategies for prompt-based semantic parsers in order to advance this field. In addition, our focus is limited to text-to-SQL tasks, and we encourage further investigation into the robustness of semantic parsers across different datasets and LFs. Despite these limitations, we emphasize the importance of exploring more effective prompt design in order to enhance the robustness of prompt-based semantic parsers, including \codex, which shows non-negotiable vulnerability.

\bibliography{custom}
\bibliographystyle{acl_natbib}
\clearpage
\appendix
\input{sections/appendix}

\end{document}

%% file: definition.tex
\usepackage[utf8]{inputenc} % allow utf-8 input
\usepackage[T1]{fontenc}    % use 8-bit T1 fonts
\usepackage{url}            % simple URL typesetting
\usepackage{booktabs}       % professional-quality tables
\usepackage{amsfonts,amsmath}       % blackboard math symbols
\usepackage{nicefrac}       % compact symbols for 1/2, etc.
\usepackage{microtype}      % microtypography
\usepackage{xcolor}
\usepackage{enumitem}
\usepackage{graphicx}
\usepackage{xspace}
\usepackage{soul}
\usepackage{multirow}
\usepackage{bbold}
\usepackage{makecell}
\usepackage{hyperref}       % hyperlinks
\usepackage{cleveref}
\usepackage{bm}
\usepackage{makecell, tabularx}
\usepackage{arydshln}
\usepackage{pifont}
\usepackage{placeins}

\newcommand{\codex}{\textsc{Codex}\xspace}
\newcommand{\davinci}{\texttt{code-davinci-002}\xspace}
\newcommand{\geo}{\texttt{GeoQuery}\xspace}

\newcommand{\scholar}{\texttt{Scholar}\xspace}

%% file: math_comm.tex
\usepackage{amsthm,amsmath,amsfonts,bm,xspace}
\usepackage{color}

\newcommand{\comment}[1]{}

\def\eqref#1{(\ref{#1})}
\def\1{\bm{1}}

\def\vx{{\bm{x}}}
\def\vy{{\bm{y}}}

\DeclareMathAlphabet{\mathsfit}{\encodingdefault}{\sfdefault}{m}{sl}
\SetMathAlphabet{\mathsfit}{bold}{\encodingdefault}{\sfdefault}{bx}{n}

\DeclareMathOperator*{\argmax}{arg\,max}

%% file: sections/abstract.tex
\begin{abstract}
Semantic parsing is a technique aimed at constructing a structured representation of the meaning of a natural-language question. Recent advances in language models trained on code have shown superior performance in generating these representations compared to language models trained solely on natural language text. The existing fine-tuned neural semantic parsers are vulnerable to adversarial attacks on natural-language inputs. While it has been established that the robustness of smaller semantic parsers can be enhanced through adversarial training, this approach is not feasible for large language models in real-world scenarios, as it requires both substantial computational resources and expensive human annotation on in-domain semantic parsing data. This paper presents the first empirical study on the adversarial robustness of a prompt-based semantic parser based on \codex, a state-of-the-art (SOTA) language model trained on code. Our results demonstrate that the large language model of code is vulnerable to carefully crafted adversarial examples. To overcome this challenge, we propose methods for enhancing robustness without requiring substantial amounts of labelled data or intensive computational resources.
\end{abstract}

%% file: sections/01-introduction.tex
\section{Introduction}

%With its powerful ability to transduce natural-language utterances (NLs) into machine-readable logical forms (LFs)~\cite{zettlemoyer2005learning}, semantic parsing has now found a wide application in numerous research areas such as code generation, question-answering systems, and dialogue systems~\cite{kamath2018survey}. Most state-of-the-art semantic parsers are deep learning models trained in a fully supervised manner on the in-domain data. However, training such models requires a high volume of in-domain semantic parsing data, which is expensive to acquire~\cite{bapna2017towards}.
Semantic parsing is a technique that transforms natural-language utterances (NLs) into machine-readable logical forms (LFs) and has been widely applied in various research fields, such as code generation, question-answering systems, and dialogue systems~\cite{kamath2018survey}. Most current state-of-the-art semantic parsers are deep-learning models trained in a supervised manner using in-domain data. However, this approach requires a large amount of in-domain semantic parsing data, which can be costly to obtain~\cite{bapna2017towards}.

To address this issue, prompt-based semantic parsers based on large pre-trained language models, such as Codex~\cite{chen2021evaluating} and GPT-J~\cite{gpt-j}, have become a new choice for semantic parsing applications. Prompt-based semantic parsers learn to solve a new task by in-context learning, instructing the parsers to generate correct LFs by constructing the prompt with a few demonstration examples. Such a method can significantly lower the cost of annotations by including only a few exemplars in the prompt and achieve comparable results to fully-supervised semantic parsers~\cite{shin-van-durme-2022-shot}.

Recent studies~\cite{huang2021robustness,pi2022towards,zhuo2023exploring} show that fully-supervised semantic parsers and language models are vulnerable to adversarial attacks, which perturb input sentences into their semantic equivalent adversaries to mislead models to produce attacker-desired outputs. 
Hence, to mitigate such attacks, various adversarial training methods~\cite{tramer2019adversarial,shafahi2019adversarial,ganin2016domain,shafahi2020universal} have been proposed to improve the adversarial robustness of the semantic parsers. In light of this, two  main questions naturally arise: (1) \textit{Do prompt-based semantic parsers based on large pre-trained language models also suffer from adversarial attacks?} (2) \textit{If so, how can we improve the robustness of the large prompt-based semantic parsers?}

To address the former question, we evaluate the prompt-based semantic parsers on several evaluation sets built by different perturbation approaches mentioned in the AdvGLUE~\cite{wang2021adversarial} dataset. 
Adopting the adversarial evaluation metrics proposed by~\citet{huang2021robustness}, it is found that the prompt-based semantic parsers are vulnerable to various types of adversarial attacks.

According to the experimental results from the first step, we perform a three-fold experiment to answer the latter questions. The first aspect of the study aims to determine if the inclusion of additional examples within the prompt during in-context learning improves the robustness of prompt-based parsers. 
This hypothesis is based on prior research that has demonstrated that the increase in the size of the training data results in an enhancement of robustness in fully-supervised models~\cite{pang2019improving}.
The second part of the study aims to determine if the integration of few-shot adversarial examples within prompts can improve the robustness of Codex. This was based on the observation that conventional adversarial training methods often include adversarial examples within the training set \cite{miyato2016adversarial,tramer2019adversarial}.
Finally, the third part of the study aims to evaluate if sampling methods other than random sampling can select more effective examples that improve the robustness of prompt-based parsers. %from~\citet{shin-van-durme-2022-shot,shin2021constrained}.

In this work, we perform a series of experiments to probe \codex, a large pre-trained model trained on code, on two semantic parsing benchmarks, \geo~\cite{zelle1996learning} and \scholar~\cite{iyer2017learning}. Our key findings from the above experiments are as follows: 
\begin{itemize}

\item Prompt-based semantic parsers are vulnerable to adversarial examples, particularly the ones crafted by sentence-level perturbations.

\item In-context learning with more demonstration examples in the prompt can improve the in-domain robustness of prompt-based parsers.

\item Augmenting the prompt with adversarial examples has limited effect in improving the robustness of prompt-based parsers.

\item The few-shot example sampling strategy with higher language complexity can result in stronger robustness for the prompt-based parsers.

\end{itemize}

\begin{figure*}
    \centering
\end{figure*}

%% file: sections/05-related-work.tex
\section{Related Work}

\paragraph{Prompt-based Learning.}

Prompt-based learning is an alternative approach to supervised learning that aims to reduce the reliance on large human-annotated datasets~\cite{liu2021pre}. Unlike traditional supervised models, which estimate the probability of an output given an input text, prompt-based learning models estimate the probability of the text directly. This is achieved by applying prompt functions to modify the input text into various prompt templates with unfilled slots. By filling these slots, various Natural Language Processing (NLP) tasks can be completed, such as common-sense reasoning~\cite{kojima2022large}, self-rationalization~\cite{marasovic2021few}, and text style transfer~\cite{suzgun2022prompt}. The development of prompt-based methods has enabled zero-shot and few-shot learning in a variety of artificial intelligence domains~\cite{ramesh2021zero,yang2022empirical,sanghi2022clip}. Recent research has also evaluated the capabilities of few-shot prompt-based learning for semantic parsing~\cite{shin-van-durme-2022-shot,BenchCLAMP,drozdov2022compositional}. Our contribution extends the current research by investigating the effect of prompts comprising only a limited number of examples on the robustness of prompt-based semantic parsers.
\paragraph{Adversarial Robustness.}

Neural networks have achieved impressive performance across various domains. However, as demonstrated by~\citet{szegedy2013intriguing}, neural models are vulnerable to adversarial examples. Adversarial attacks in NLP normally take on various forms, including character-level manipulations~\cite{hosseini2017deceiving,ebrahimi2018hotflip,belinkov2018synthetic,gao2018black,eger2019text,boucher2022bad}, sentence-level rewriting~\cite{iyyer2018adversarial,ribeiro2018semantically,zhao2018generating}, and adversarial word substitutions~\cite{alzantot2018generating, liang2018deep, zhang2019theoretically}.

There has been an increasing interest in defending against adversarial attacks in large language models via adversarial training~\cite{yi2021improved,ross2022tailor,bartolo2021improving,guo2021gradient}. Adversarial training involves incorporating adversarial examples in the training set, thus making the model robust to such attacks. However, adversarial training can sometimes negatively impact the generalization ability of the neural models~\cite{raghunathan2019adversarial,min2021curious}.

%% file: sections/02-attack-method.tex
\section{Robustness Evaluation for Prompt-based Semantic Parsing}
\label{sec:2}
This section gives an overview of our evaluation framework, including the methods of constructing the evaluation corpora and the evaluation metrics to evaluate the robustness of the prompt-based semantic parser.

\input{sections/tables/adversarial_examples}
\subsection{Construction of the Evaluation Corpus}

A robust prompt-based semantic parser should be able to parse both the utterances and their adversarial counterparts into correct LFs. As proposed by~\citet{huang2021robustness}, an adversary of an utterance for a semantic parser is defined as i) an utterance with the same semantic meanings as the original one given the human judgment and ii) an utterance on which the semantic parser cannot produce correct LF. Therefore, to evaluate the robustness of prompt-based semantic parsers, we craft the robustness evaluation sets by perturbing the original utterances in existing benchmark datasets with multiple adversarial perturbation methods. Such perturbations should not alter the semantics of the original utterances. Each example in a robustness evaluation set is a perturbed utterance paired with its ground-truth LF. Next, we introduce the details of each perturbation method and how we guarantee the perturbations do not change the semantics. Table~\ref{tab:examples} illustrates some meaning-preserved utterances after perturbation in the robustness evaluation set of \geo based on different perturbation methods. More examples can be found in Appendix~\ref{app:more_examples}.

\subsubsection{Adversarial Perturbations}
\label{sec:perturb}
Following the principles as in~\citet{wang2021adversarial} to design adversarial attacks, we perform five word-level perturbations and two sentence-level perturbations to generate seven robustness evaluation sets for the standard evaluation set in each benchmark.

\paragraph{Word-level Perturbations.}
\begin{itemize}
\item \textbf{Typo-based (TB)} uses TextBugger~\cite{li2018textbugger} to replace two words in each utterance with the typos.
\item \textbf{Random Deletion (RD)} randomly deletes two words in the utterance.
\item \textbf{Random Swap (RS)} swaps the positions of two random words in each utterance.
\item \textbf{Context-aware Substitution (CS)} leverages RoBERTa~\cite{liu2019roberta} to substitute two random words with their synonyms.
\item \textbf{Context-aware Insertion (CI)} inserts two most probable words selected by RoBERTa at two random positions in each utterance.
\end{itemize}

\paragraph{Sentence-level Perturbations.}

\begin{itemize}
\item \textbf{Rewriting-based (RB)} chooses Quillbot\footnote{\url{https://www.quillbot.com/paraphrasing}}~\cite{fitria2021quillbot}, a state-of-the-art (SOTA) commercial paraphrasing model, to rewrite the complete utterances. Quillbot has been demonstrated as an effective tool to paraphrase utterances in semantic parsing data~\cite{shiri2022paraphrasing}.
\item \textbf{Distraction-based (DB)} appends interrogation statements to the end of each NL, inspired by StressTest~\cite{naik2018stress}. Specifically, we design the following interrogation statements: "who is who; what is what; when is when; which is which; where is where", in which the selected interrogative words are more likely to appear in the utterance. 

\end{itemize}

\subsubsection{Data Filtering}
\label{sec:curation}
In order to ensure that the perturbed examples preserve the meaning of the original NL, we design a two-stage evaluation process:

\textbf{Step1:} We first generate 20 adversarial examples against the original NL for each perturbation method and choose the top 10 candidates ranked based on text similarity scores between the original and the perturbed ones, which are calculated by Sentence-BERT~\cite{reimers2019sentence}.

\textbf{Step2:} We engage human experts to select the best one among the 10 adversarial candidates produced in \textbf{Step1}.% We acquire human experts' engagement to further select the best one among 10 adversarial candidates obtained from \textbf{Step1}.

\subsection{Evaluation Metrics}

Since the output LFs of the prompt-based language models may not follow the same naming convention~\cite{shin2021constrained,shin-van-durme-2022-shot} as the ground truth, previous string-based evaluation metrics, including BLEU~\cite{papineni2002bleu} and Exact Match~\cite{poon2009unsupervised}, are not suitable for prompt-based semantic parsers. Therefore, we follow~\citet{rajkumar2022evaluating} to report the \textit{execution accuracy}, which is based solely on the execution correctness of the LFs on the test sets, for the purpose of robustness evaluation.

Following \citet{huang2021robustness}, we report the experiment results with three variants of execution accuracy, namely \textit{standard accuracy}, \textit{perturbation accuracy} and \textit{robust accuracy}: 
\begin{itemize}
    \item \textbf{Standard Accuracy} is measured on the standard (original) test sets.
    \item \textbf{Perturbation Accuracy} tests the performance of the model on perturbed test sets.
    \item \textbf{Robust Accuracy} is defined as $n/|R_{eval}|$. $R_{eval}$ denotes a subset of the perturbed test sets, and $n$ is the number of the utterances in $R_{eval}$ that are parsed correctly. More specifically, $R_{eval}$ consists of the examples whose counterparts before perturbation are parsed correctly. Intuitively, Robust Accuracy estimates the quantity of cases that a parser can successfully parse before perturbation but cannot do so after perturbation, and hence shows the robustness of the parsers against adversarial perturbation.
\end{itemize}

% More specifically, w

%% file: sections/tables/adversarial_examples.tex
\begin{table*}[t]\small \setlength{\tabcolsep}{7pt}
\centering

% \resizebox{1.0\textwidth}{!}{
\begin{tabular}{p{2.0cm}p{9cm}}
\toprule 
Linguistic \quad Phenomenon & Samples (\st{Strikethrough} = Original Text, \textcolor{red}{\textbf{red}} = Adversarial Perturbation) \\
\midrule
 \multirow{2}{*}{\shortstack{Typo \\ (Word-level)}} & \multirow{2}{*}{\textbf{NL}: what can you \st{tell}\textcolor{red}{\textbf{te11}} me about \st{the}\textcolor{red}{\textbf{th e}}  population of missouri} \\\\
  \midrule
 \multirow{2}{*}{\shortstack{Substitution\\ (Word-level)}} & \multirow{2}{*}{\textbf{NL}: what \st{can}\textcolor{red}{\textbf{will}} you tell me about \st{the}\textcolor{red}{\textbf{a}} population of missouri} \\\\
  \midrule
 \multirow{2}{*}{\shortstack{Paraphrase \\ (Sent.-level)}} & \multirow{1}{*}{\textbf{NL}: \st{what can you tell me about the population of missouri}}  \\ & \textcolor{red}{\textbf{What information can you provide on Missouri's population?}}\\
\bottomrule
\end{tabular}
% }
\caption{\small \textbf{Examples from Robustness Evaluation Set}. We show 3 examples from \geo. These examples are generated with three different perturbations, and they all can successfully change the predictions of \codex.}

 \label{tab:examples}
\end{table*}

%% file: sections/03-weakly-adversarial-guidence.tex
\section{Improving Robustness of Prompt-based Semantic Parsers}

Instead of predicting the LF conditioned on the input utterance, large language models such as \codex could learn to solve a specific task by \textit{in-context learning}. During in-context learning, the parser predicts the LF conditioned on a prompt which consists of a small list of utterance-LF pairs to demonstrate the semantic parsing task and, optionally, a table schema. To defend against adversarial attacks, one seminal approach is adversarial training. One of the most typical adversarial training methods augments the training data with adversarial examples, from which the machine learning model could learn robust features~\cite{allen2022feature} by gradient descent. However, directly adapting conventional adversarial training is not suitable for in-context learning. First, the number of demonstration examples is limited due to the restriction on the maximum number of tokens for the pre-trained language model. As a result, we cannot include an arbitrary number of adversarial examples in the prompt, which might not include enough robust features. Second, in-context learning does not update the parameters of the language model. The model would not be optimized towards learning the robust features in the adversarial examples through gradient descent.

%Prior works~\cite{brown2020language, schick2021s, madotto2020language} suggest that in-context few-shot learning is sufficient for prompt-based language models to achieve SOTA performance in various NLP tasks. However, there is no existing study on whether in-context few-shot learning can improve the robustness of these language models. 

Given the difference, it is unclear whether in-context learning could improve the robustness of the parser as the conventional supervised training. In this paper, we conduct the first investigation on in-context learning for model robustness. More specifically, we examine the impact of variants of in-context learning and sampling methods on parser robustness.

\subsection{Standard In-context Few-shot Learning}
In our setting, given an input utterance $\vx$, the pre-trained language model $P(\cdot;\theta)$ predicts the LF $\vy'$ conditioned on the prompt, which consists of a set of demonstration examples $\mathcal{M} = \{(\vx_i,\vy_i)\}^{N}_{i=1}$, and a table schema $\mathcal{T}$:
\begin{equation}
    \vy' = \argmax_{\vy \in \mathcal{Y}} P(\vy|\vx, \mathcal{M}, \mathcal{T};\theta)
\end{equation}
For the \textit{few-shot} setting, the number of demonstration examples $N$ is limited by a budget size.

\subsection{Adversarial In-context Few-shot Learning}
In adversarial in-context learning, we include the perturbed adversarial examples, $\mathcal{M}_{adv}$, in the demonstration examples:
\begin{equation}
    \vy' = \argmax_{\vy \in \mathcal{Y}} P(\vy|\vx, \mathcal{M}\cup \mathcal{M}_{adv}, \mathcal{T};\theta)
\end{equation}

\subsection{In-context Few-shot Selection}
\label{sec:sampling}
Current in-context learning assumes there is an example pool from where they can select prompting examples. However, most of the works only randomly pick examples from the pools. We argue that the way to select the examples might deeply impact the robustness of the prompt-based semantic parser. Therefore, we examine various strategies to select in-context few-shot examples. 

\paragraph{Random Sampling (Random).} We randomly sample $N$ utterances from the example pool.
\paragraph{Confidence-based Sampling (Confidence)~\cite{duong2018active}.} We score each utterance with the confidence of the parser on the predicted LF given the utterance and the table schema. Then we select the ones with the lowest parser confidence scores\footnote{The confidence scoring parser is a zero-shot model, meaning that there are no examples present in the prompt. It operates solely based on the input utterance, instructions, and schema provided within the prompt. Please see Section~\ref{sec:setup} for more information on the prompt structure.}.
\paragraph{Diversity-based Sampling.} Following~\citet{li2021total}, we partition the utterances in the utterance pool into $N$ clusters with the K-means~\cite{wu2012advances} clustering algorithm and select the example closest to the cluster centers. We measure the edit distance (Cluster-ED)~\cite{wagner1974string}, and Euclidean distances using utterance features of TF-IDF (Cluster-TF-IDF)~\cite{anand2011mining}, or Contextual Word Embedding (Cluster-CWE) encoded by Sentence-BERT~\cite{reimers2019sentence}, between each pair of utterances for K-means.
\paragraph{Perplexity-based Sampling~\cite{sen-yilmaz-2020-uncertainty}.} We score each utterance with the perplexity of GPT-2 on this utterance. Then we select the utterances with the highest (PPL. Asc) and lowest (PPL. Desc) perplexity scores, respectively. 

%% file: sections/04-experiment.tex
\section{Experiments}
\input{sections/tables/rq_1}
\subsection{Setup}
\label{sec:setup}
\paragraph{Evaluation Datasets.} We evaluate the robustness of the prompt-based semantic parsers via the adversarial robustness sets built on top of the test sets of \geo~\cite{finegan2018improving} and \scholar~\cite{finegan2018improving} with the proposed perturbation methods in Section~\ref{sec:2}. As in~\citet{finegan2018improving}, we choose the \textit{query} splits of both \geo and \scholar, where there is no LF template overlap among train, test, and dev sets.

\paragraph{Prompt-based Semantic Parser.} We choose \codex~\cite{chen2021evaluating} as the representative prompt-based semantic parser for our evaluation. In recent studies, \codex has performed comparably via in-context few-shot semantic parsing to the SOTA-supervised trained neural semantic parsers~\cite{shin-van-durme-2022-shot, roy2022benchclamp, drozdov2022compositional} in terms of execution accuracy.

To examine the vulnerability of large prompt-based semantic parsers against adversarial examples, we choose the \davinci version of \codex as it is the most powerful variant among all \codex models, with 175B parameters. In our experiments, we sample a maximum of 200 tokens from \codex with the temperature set to 0, with the stop token to halt generation.

\paragraph{Prompts.} %We use the prompt design of \textbf{Create Table + Select X}~\cite{rajkumar2022evaluating}, which proves the effectiveness of \textbf{static prompting}~\footnote{We use the term ``static prompting'' for semantic parsing in contrast to dynamically selected few-shot examples from the example pool as in the work of \citet{shin2021constrained}. In this work, we refer `in-context few-shot learning' to the few-shot examples which do not change in context.} for semantic parsing on \codex. The prompt provides \texttt{CREATE TABLE} commands for each table of the dataset, including column type, foreign key declarations, and in comments the results of executing a \texttt{SELECT * FROM T LIMIT X} query on each table with schemas via column headers. Subsequently, we apply the strategies introduced in Section~\ref{sec:sampling} to select NL-LF pairs as in-context few-shot examples from $train$ sets. To guide the prompt-based semantic parser, we append the same textual instruction of ``\texttt{Using valid SQLite, answer the following questions for the tables provided above.}'' as the one in \citet{rajkumar2022evaluating}.

In this work, we adopt the prompt design of \texttt{Create Table + Select X} as presented in~\citet{rajkumar2022evaluating}, which has been shown to be effective for semantic parsing using \textit{static prompting}\footnote{In contrast to the approach of~\citet{shin2021constrained} which involves dynamically selecting few-shot examples from an example pool, we refer to static prompting as being performed with a fixed set of examples.}.

The prompt for semantic parsing on \codex consists of \texttt{CREATE TABLE} commands, including specifications for each table's columns, foreign key declarations, and the results of executing a \texttt{SELECT * FROM T LIMIT X} query on the tables via the column headers. As described in Section~\ref{sec:sampling}, we select NL-LF pairs as in-context few-shot examples from the $train$ sets.

To guide the prompt-based semantic parser, we also include the textual instruction of ``\texttt{Using valid SQLite, answer the following questions for the tables provided above.}'' as proposed by~\citet{rajkumar2022evaluating}.

\subsection{Research Questions and Discussions}
Our experimental results answer the following four research questions (RQs) related to the robustness of \codex.

\subsubsection*{RQ1: How vulnerable is the prompt-based semantic parser to adversarial examples?}

\paragraph{Settings.} To answer RQ1, we evaluate the standard accuracy and perturbation accuracy of \codex on \geo and \scholar test sets through \emph{zero-shot} learning.

\paragraph{Results.} 

The zero-shot parsing performances of \codex are shown in Table~\ref{tab:vulnerability}. Our first observation is that \codex is more vulnerable to sentence-level perturbations than to word-level perturbations, as indicated by the more significant performance gaps between standard and perturbed accuracies on the sentence-level perturbed test sets. \citet{wang2021adversarial} observed that neural language models are vulnerable to human-crafted adversarial examples where there are complex linguistic phenomenons (e.g., coreference resolution, numerical reasoning, negation). We observe that the rewriting model trained on human paraphrase pairs also introduces such complex linguistic phenomenons. 

With respect to the word-level perturbations, \codex is most robust to typo-based perturbations, which is surprising as \citet{wang2021adversarial} shows typo-based perturbation is the most effective attack method for large language models like BERT~\cite{devlin2019bert} in the evaluation of natural language understanding tasks. However, utterances with typos drop only 3\% of the accuracy of \codex. Random Deletion is also less effective than the other word-level methods, consistent with the observations by~\citet{huang2021robustness} on the fully-supervised semantic parsers. This phenomenon can be attributed to the fact that Random Deletion primarily makes minor modifications to the standard NL utterances, as this method often involves removing non-functional words such as articles, for example, ``the'' and ``a.''

Although \codex is pre-trained on a considerably large dataset, it does not show robustness on the in-domain tasks. We conjecture that the reason is that zero-shot \codex has not yet learned any in-domain knowledge on \geo or \scholar. So in RQ2, we would address whether in-domain examples would improve the robustness of \codex.

\paragraph{Takeaways.} Zero-shot \codex is vulnerable to adversarial examples, especially sentence-level perturbation of utterances, rather than to word-level perturbations.

\input{sections/tables/rq_2}
\subsubsection*{RQ2: Does standard in-context few-shot learning improve the robustness?}
\paragraph{Settings.}

We respectively select up to 50 and 10 examples from \geo and \scholar train sets\footnote{Due to the larger size of table schema in \scholar, we could only include up to 10 examples in the prompt.}, with the random sampling strategy, to construct prompts for parsers. Then, for each few-shot learning experiment, we measure standard accuracy, perturbation accuracy and robust accuracy on our various perturbed test sets.

\paragraph{Results.} Tables~\ref{tab:geo_few_shot} and \ref{tab:scholar_few_shot} show the performance of standard in-context few-shot learning on the robustness evaluation sets perturbed by different methods. We observe that more standard examples in the prompt can evidently improve the robust accuracy of \codex, which demonstrates the effectiveness of standard in-context few-shot learning in improving the robustness of semantic parsing. Although it performs slightly worse on the test sets perturbed by typo-based methods than the one perturbed by the random deletion in \geo, we argue that this is due to the performance variance, which does not necessarily hurt the model robustness.

The performance gap between perturbation accuracy and standard accuracy is enlarged as the number of in-context shots increases from 5 to 20, and then remains stable. However, the robust accuracy grows slowly. This indicates that improving the generalization ability of the parser does not necessarily mean the improvement of the robustness. The trade-off between standard and robust accuracies is a long-standing problem in adversarial training. \citet{raghunathan2019adversarial} shows that increasing the training sample size could eliminate such a trade-off. Our experiments demonstrate that in-context learning follows similar patterns as supervised adversarial training. It can be observed that both objectives can be improved with a limited number of examples when compared to the zero-shot parser. However, the extent of improvement varies.

\paragraph{Takeaways.} With more standard in-context examples, few-shot \codex can be guided to achieve better robustness and standard execution performance.

\input{sections/tables/rq_3}

\subsubsection*{RQ3: Does adversarial in-context learning improve robustness?} 

\paragraph{Settings.} In this work, we present the experimental results of \codex on both \geo and \scholar datasets, using 10 and 5 in-context examples, respectively. In order to assess the robustness of \codex through adversarial in-context learning, we first augment the standard few-shot examples by incorporating examples whose utterances have been perturbed using various perturbation methods. Subsequently, for each set of augmented examples, we calculate the average robust accuracy of \codex based on the average of the parser robust accuracies on all robustness evaluation sets.

\paragraph{Results.} %Table~\ref{tab:overall_robustness} summarizes the results when we vary perturbation strategies for the in-context few-shot examples. Supervised adversarial training has always been considered an effective approach to improving the robustness of machine learning models. However, on both \geo and \scholar, the robust accuracies are only marginally improved by adversarial in-context learning. In addition, many studies~\cite{raghunathan2019adversarial,huang2021robustness} point out supervised adversarial training sometimes hurts standard accuracy while it improves robust accuracy. However, adversarial in-context learning shows contradictory results where the standard accuracy is improved greatly on \scholar while the robust accuracy is only marginally improved. The two observations show that adversarial in-context learning differs from supervised adversarial training in improving the model's robustness. In contrast to supervised adversarial training, simply augmenting the prompt with adversarial examples has a limited effect on the robustness of parsers.

The experimental results of the various perturbation strategies applied to the in-context few-shot examples are presented in Table~\ref{tab:overall_robustness}. While the approach of supervised adversarial training has been widely regarded as an effective means of enhancing the robustness of machine learning models, the results indicate that on both \geo and \scholar, the robust accuracies are only marginally improved through the application of adversarial in-context learning. Previous studies~\cite{raghunathan2019adversarial,huang2021robustness} have pointed out that supervised adversarial training can sometimes result in a decrease in standard accuracy, even as it improves robust accuracy. However, the results of adversarial in-context learning diverge from this trend, with significant improvement in standard accuracy, from 23\% to more than 33\%, observed on \scholar, while robust accuracy only experiences marginal improvement. These observations indicate that adversarial in-context learning represents a distinct approach from supervised adversarial training in terms of enhancing the robustness of the model. Furthermore, the results suggest that simply incorporating adversarial examples into the prompt has a limited impact on the robustness of parsers, in contrast to supervised adversarial training.
\input{sections/tables/rq_4}

\input{sections/tables/ablation}
Of all the perturbation strategies analyzed, the results indicate that \codex achieves the best performance in terms of both standard and robust accuracy through the application of RB adversarial in-context learning, but experiences the worst performance through TB adversarial in-context learning. The hypothesis is that RB produces utterances with more complex linguistic features, resulting in enhanced standard and robust accuracy during in-context learning. To test this hypothesis, the number of standard examples is doubled (No Adv.$\times$2) to match the size of the examples augmented with the adversarial examples. The results show that the robust and standard accuracies of \codex are higher than those obtained through adversarial in-context learning, likely due to the greater diversity of linguistic variations in the doubled standard examples.

\paragraph{Takeaways.} The robustness of few-shot \codex can be marginally improved by adversarial in-context learning without significant negative impacts on standard performances.

\subsubsection*{RQ4: What is the impact of sampling techniques on the robustness of parsers that utilize standard in-context few-shot learning?}
\label{rq3}
\paragraph{Settings.} In order to compare the influence of different sampling methods on the robustness of the model, we vary standard in-context examples on \geo and \scholar with all 7 strategies aforementioned in Section \ref{sec:sampling}. We choose the 50-shot setting for \geo and 10-shot setting for \scholar.

\paragraph{Results and Takeaways.} We present standard accuracies in Table~\ref{tab:context_robustness} when varying the sampling methods for the few-shot example selection. We first observe that different sampling strategies impact the robust and standard performance of the \codex. Overall, the Cluster methods, which diversify the features of the selected utterances, perform better than the other sampling methods. On the other hand, PPL. Desc sampling method performs consistently poorly than the other sampling methods. In brief, we conclude that \codex is sensitive to the few-shot example sampling strategies. 

\subsection*{RQ4-Ablation: Why does \codex react differently to various sampling strategies?}

\paragraph{Settings.} 
The findings of RQ 1 and RQ 3 indicate that linguistic complexity has an impact on the performance of \codex. As a result, the results of RQ 4 may also be influenced by linguistic complexity. To further explore this correlation, three lexical diversity metrics, Type-Token Ratio (TTR)~\cite{templin1957certain}, Yule's I~\cite{yule1944statistical}, and Measure of Textual Lexical Diversity (MTLD)~\cite{mccarthy2005assessment}, are adopted to measure the lexical diversity of the selected NLs from \geo and \scholar. 
The TTR is defined as the ratio of the number of unique tokens, also known as types, to the total number of tokens. The TTR is used as an indicator of lexical diversity, with a higher TTR indicating higher lexical diversity. Yule's Characteristic Constant (Yule's K) is a measure of text consistency that considers vocabulary repetition. Yule's K and its inverse, Yule's I, are considered more robust to variations in text length than the Type-Token Ratio (TTR). MTLD is computed as the average number of words in a text required to maintain a specified TTR value.

%Yule's characteristic constant (Yule's K) measures of text consistency include vocabulary repetition. Yule's K and its inverse Yule's I are considered to be more resistant to variations in text length than TTR~\cite{oakes2012}. MTLD is evaluated sequentially as the average number of words in a text that retain a certain TTR value~\cite{mccarthy2005assessment}.

%The specifics of these three lexical diversity metrics are explained in Appendix~\ref{app:diversity}.

\paragraph{Results and Takeaways.} Table~\ref{tab:complexity} presents the lexical diversity of each set of NLs sampled by different approaches. The diversity scores obtained from the three metrics align with the performance of \codex as presented in Table~\ref{tab:context_robustness}. For instance, the three metrics indicate that the examples selected using the Cluster-TF-IDF method achieve higher lexical diversity compared to those selected through the other methods. Additionally, the Cluster-TF-IDF method also produces the highest results in terms of robust and standard accuracy for \codex. Thus, it can be inferred that an increase in the lexical diversity of the few-shot examples leads to an improvement in the robust and standard accuracy of \codex.

%% file: sections/tables/rq_1.tex
\begin{table*}[]
    \small
    \centering
    \begin{tabular}{cc||c|c|c||c|c|c}
    \hline
    \hline
        \multirow{2}{*}{Category} & \multirow{2}{*}{Pert. Strategy} & \multicolumn{3}{c||}{\geo} & \multicolumn{3}{c}{\scholar}\\\cline{3-8}
        & & Pert. Acc. & Std. Acc. & $\Delta$  & Pert. Acc. & Std. Acc. & $\Delta$ \\\hline
        \multirow{5}{*}{Word-level}
        & TB & 53.85 & \multirow{7}{*}{57.14}& -3.29& 11.35& \multirow{7}{*}{12.21} &-0.86 \\\cline{3-3}\cline{5-6}\cline{8-8}
        & RD & 50.55 & &-6.59 & 10.52& &-1.69 \\\cline{3-3}\cline{5-6}\cline{8-8}
        & RS & 37.36 & & -19.78& 8.31 & &-3.90 \\\cline{3-3}\cline{5-6}\cline{8-8}
        & CS & 42.31 & & -14.83& 8.40& &-3.81 \\\cline{3-3}\cline{5-6}\cline{8-8}
        & CI & 38.46 & &-18.68 & 8.31 & &-3.90 \\\cline{3-3}\cline{5-6}\cline{8-8}
        \cline{1-3}\cline{5-6}\cline{8-8}
        \multirow{2}{*}{Sentence-level}
        & RB & \textbf{31.87} & &-25.27 & \textbf{5.22} & & -6.99 \\\cline{3-3}\cline{5-6}\cline{8-8}
        & DB & 35.71 & &-21.43 & 7.88& & -4.33 \\\bottomrule
        \hline
         \hline
    \end{tabular}
    \caption{Results of perturbation accuracy (Pert. Acc.) and standard accuracy (Std. Acc.) of zero-shot performance on \geo and \scholar. The zero-shot prompt only contain the table information and initial semantic parsing instruction. Perturbation accuracy is calculated based on each perturbation method.}
    \label{tab:vulnerability}
\end{table*}

%% file: sections/tables/rq_2.tex
% \begin{table}[]
%     \small
%     \centering
%     \begin{tabular}{cc|cc|c}
%          \hline
%          \multirow{1}{*}{Dataset} & $n$-shot & Pert. & Std. & Robust\\\hline
%          \multirow{3}{*}{\geo} & 10 & 56.80 & 74.73 & 47.28\\
%          & 20 & 57.50 & 79.67 & 48.74\\
%          & 50 & 66.58 & 84.62 & 53.22\\\hline
%          \multirow{3}{*}{\scholar} & 3 & 7.71 & 31.12 & 14.30 \\
%          & 5 & 16.27 & 33.10 & 23.08\\
%          & 10 & 25.35 & 35.97 & 40.32\\\hline

%     \end{tabular}
%     \caption{\TZ{To be done}}
%     \label{tab:in_context}
% \end{table}

\begin{table}[ht]
    \small
    \centering
\resizebox{1.0\linewidth}{!}{
    \begin{tabular}{c|c|c|c|c|c|c}
    \hline
    \hline
         \multirow{1}{*}{Pert. Strategy} 
         & \multicolumn{1}{c|}{5-shot}
        & \multicolumn{1}{c|}{10-shot} & \multicolumn{1}{c|}{20-shot}
        & \multicolumn{1}{c|}{30-shot} & \multicolumn{1}{c|}{40-shot}
        & \multicolumn{1}{c}{50-shot} \\\hline
        TB  & 63.19 & 71.98 & 78.02 & 81.32 & 81.30 & 82.42\\
        RD  & 59.34 & 64.29 & 71.43 & 71.98 & 70.33 & 75.27\\
        RS  & 52.20 & 52.75 & 54.87 & 59.34 & 60.99 & 63.14\\
        CS  & 54.40 & 56.04 & 60.44 & 63.19 & 65.93 & 67.03\\
        CI  & 51.65 & 54.95 & 55.49 & 57.69 & 58.79 & 61.02\\
        RB  & 44.51 & 47.80 & 49.45 & 50.55 & 54.23 & 57.27\\
        DB  & 48.35 & 49.77 & 53.30 & 54.20 & 59.34 & 59.89 \\\hdashline 
        Avg. Pert. Acc.  & 53.38 & 56.80 & 60.43 & 62.61 & 64.42 & 66.58\\\hline
        Std. Acc.  & 66.48 & 74.37 & 79.67 & 81.87 & 83.52 & 84.62\\\hline
         Avg. Robust Acc.  & 75.67 & 77.28 & 78.74 & 80.44 & 82.07 & 83.22\\\bottomrule
        \hline
         \hline

    \end{tabular}}
    \caption{Few-shot performances on \geo. We conduct \{5, 10, 20, 30, 40, 50\}-shot learning experiments. Average perturbation accuracy (Avg. Pert. Acc.) is the average score of execution accuracies on different perturbation sets. Average robust accuracy (Avg. Robust Acc.) is the average score of execution accuracies on the test sets perturbed by different perturbation methods.}
    \label{tab:geo_few_shot}
\end{table}

\begin{table}[ht]
    \small
    \centering
    \begin{tabular}{c|c|c|c}
    \hline
    \hline
         \multirow{1}{*}{Pert. Strategy} 
        & \multicolumn{1}{c|}{3-shot} & \multicolumn{1}{c|}{5-shot}
        & \multicolumn{1}{c}{10-shot}\\\hline
        TB & 10.57 & 20.33 & 34.29\\
        RD & 12.09 & 25.27 & 31.43\\
        RS & 6.04 & 17.03 & 25.08\\
        CS & 9.34 & 18.13 & 26.03\\
        CI & 8.24 & 14.29 & 20.63\\
        RB & 3.30 & 8.43 & 18.10\\
        DB & 4.40 & 10.44 & 21.90\\ \hdashline
        Avg. Pert. Acc. & 7.71 &16.27 & 25.35\\\hline
        Std. Acc. & 14.29 & 23.08 & 40.32\\\hline
        Avg. Robust. Acc. & 51.12 & 53.10 & 55.97\\\bottomrule
        \hline
         \hline

    \end{tabular}
    \caption{Few-shot performances on \scholar. We conduct \{3, 5, 10\}-shot learning experiments.}
    \label{tab:scholar_few_shot}
\end{table}

%% file: sections/tables/rq_3.tex
% \begin{table}[]
%     \small
%     \centering

%     \begin{tabular}{c|cc|cc}
%     \hline
%          \multirow{2}{*}{Pert. Strategy} & \multicolumn{2}{c|}{\advgeo} 
%          & \multicolumn{2}{c}{\advscholar}\\ \cline{2-5}
%         & \multicolumn{1}{c}{10-shot} 
%         & \multicolumn{1}{c|}{50-shot} & \multicolumn{1}{c}{5-shot} 
%         & \multicolumn{1}{c}{10-shot} \\\hline
%         TB & 35.87 &35.98 & 13.67 & 14.09\\
%         RD & 36.38 & 35.98 & 14.71 & 14.13\\
%         RS & 35.16 & 36.37 & 14.92 & 13.07\\
%         CS & 36.00 & 35.96 & 14.81 &14.08\\
%         CI & 35.49 & 36.62 & 13.46 & 13.25\\
%         QB & 35.35 & 35.80 & 14.92 & 13.57\\
%         DLS &  35.93 &35.69 & 13.25 &14.91\\ \hline

%     \end{tabular}
%     \caption{Results of the overall robust accuracy with different perturbed few-shot examples. We choose \{10, 50\}-shot on \advgeo and \{5, 10\}-shot on \advscholar. Note that we only consider adversarial examples as the few-shot context for learning in this experiment.}
%     \label{tab:overall_robustness}
% \end{table}

\begin{table}[]
    \small
    \centering
\resizebox{1.0\linewidth}{!}{
    \begin{tabular}{c|cc|cc}
    \hline
    \hline
         \multirow{2}{*}{Adv. L. Strategy} & \multicolumn{2}{c|}{\geo} 
         & \multicolumn{2}{c}{\scholar}\\ 
        & \multicolumn{1}{c}{Avg. Robust Acc.} 
        & \multicolumn{1}{c|}{Std. Acc.} & \multicolumn{1}{c}{Avg. Robust Acc.} 
        & \multicolumn{1}{c}{Std. Acc.} \\\hline
                No Adv. & 77.28 &74.37 & 53.10 & 23.08 \\
                No Adv. ($\times$ 2) & 78.74 &79.67 & 55.97 & 40.32\\
                \hline
        TB & 77.32 & 73.62 & 52.75 & 34.99\\
        RD & 77.40 & 73.64 & 53.11 & 33.65\\
        RS & 78.30 & 74.73 & 54.88 & 33.46\\
        CS & 78.05 & 74.47 & 53.12 & 34.86\\
        CI & 78.14 & 74.81 & 54.66 & 33.65\\
        % \hline
        RB & 78.47 & 75.51 & 56.58 & 35.85\\
        DB& 78.31 & 75.08 & 55.08 & 35.71\\ \bottomrule
        \hline
         \hline

    \end{tabular}}
    \caption{The results of the average robust accuracy obtained through Adversarial In-context Learning (Adv. L. Strategy) with different types of perturbed few-shot examples. Additionally, we include results of applying the method with only standard examples (No Adv.), as well as with a doubled number of standard examples (No Adv. ($\times$ 2)).}
    \label{tab:overall_robustness}
\end{table}

%% file: sections/tables/rq_4.tex
\begin{table*}[ht]
    \small
    \centering
    \resizebox{0.9\textwidth}{!}{
    \begin{tabular}{cc|cccccc|c}
    \hline
    \hline
         Dataset & Metric & Confidence & Cluster-CWE & Cluster-ED & Cluster-TF-IDF & PPL. Asc & PPL. Desc & Random\\\hline
        \multirow{3}{*}{GeoQuery} & \multirow{1}{*}{Avg. Robust  Acc.}  & 78.77 & 78.02 & 82.40 & \textbf{85.10} & 70.36 & 50.82 & 73.22\\

        & \multirow{1}{*}{Avg. Pert. Acc.}  & 69.80 & 68.93 & 74.41 & \textbf{77.74} & 62.11 & 45.71 & 66.58\\
        & \multirow{1}{*}{Std. Acc.}  & 74.73 & 78.41 & 81.32 & \textbf{85.64} & 73.14 & 70.76 & 74.18\\
        \hline
        \multirow{3}{*}{Scholar} 
        & \multirow{1}{*}{Avg. Robust  Acc.}  & 55.31 & 60.24 & 60.68& \textbf{62.61}& 53.97& 47.18& 55.97\\

        & \multirow{1}{*}{Avg. Pert. Acc.}  & 28.55  & 29.81 & 30.28 & \textbf{31.44} & 22.25 & 7.47 & 25.35\\

        & \multirow{1}{*}{Std. Acc.}  & 37.99 & 41.49 & 42.19 & \textbf{42.97} & 35.93 & 34.89 & 36.91\\
        \bottomrule
        \hline
         \hline
    \end{tabular}
    }
    \caption{The performance of standard few-shot in-context learning using various sampling methods on the \geo and \scholar datasets. The average robust accuracy, average perturbation accuracy, and standard accuracy are computed for each sampling method to assess their efficiency in this learning scenario. The Random column comes from a separate random-sampling run and therefore differs from the corresponding settings in Tables~\ref{tab:geo_few_shot} and \ref{tab:scholar_few_shot}.}
    \label{tab:context_robustness}
\end{table*}

%% file: sections/tables/ablation.tex
\begin{table*}[hbt!]
    \small
    \centering
    \resizebox{1.0\linewidth}{!}{
    \begin{tabular}{cc|ccccccc}
    \hline
    \hline
         Dataset & LC. Metric & Confidence & Cluster-CWE & Cluster-ED & Cluster-TF-IDF & PPL. Asc & PPL. Desc & Random\\\hline
        \multirow{3}{*}{GeoQuery} & TTR $\uparrow$ & 7.68 & 7.24 & 8.47 & 
        \textbf{10.26} & 5.94 & 3.22 & 6.44\\
        & Yule's I $\uparrow$& 68.55 & 64.37 & 69.59 & \textbf{71.49} & 62.94 & 43.41 & 58.14\\
        & MTLD $\uparrow$& 12.44 & 12.19 & 13.37 & \textbf{15.58} & 11.32 & 8.16 & 10.41\\\hline
        \multirow{3}{*}{Scholar} & TTR $\uparrow$ & 28.18 & 29.91 & 31.40 & \textbf{33.11} & 21.15 & 14.17 & 25.67\\
        & Yule's I $\uparrow$ & 198.15 & 207.11 & 223.76 & \textbf{262.36} & 177.37 & 102.17 & 193.31\\
        & MTLD $\uparrow$ & 15.68 & 15.63 & 16.34 & \textbf{19.49} & 11.94 & 13.12 & 14.36\\\bottomrule
        \hline
         \hline
    \end{tabular}}
    \caption{Results of the language complexity of standard NLs sampled by different sampling strategies, measured by three lexical diversity (LC.) metrics. For the ease of readability and comparison, we multiply both TTR scores and Yule’s I scores by 100.
}
    \label{tab:complexity}
\end{table*}

%% file: sections/06-conclusion.tex
\section{Conclusion}
This study examines the robustness of semantic parsers in the context of prompt-based few-shot learning. To achieve this objective, robustness evaluation sets were curated to evaluate the robustness of the prompt-based semantic parser, \codex. The research aims to identify methods to improve the robustness of \codex. The results of our comprehensive experiments demonstrate that even the prompt-based semantic parser based on a large pre-trained language model is susceptible to adversarial attacks. Our findings also indicate that various forms of in-context learning can improve the robustness of the prompt-based semantic parser. It is believed that this research will serve as a catalyst for future studies on the robustness of prompt-based semantic parsing based on large language models.

%% file: sections/appendix.tex
\onecolumn
\section{Experiments}
%\geo is proposed based on a database consisting of U.S. geographical information, and natural language questions, which contains 877 NL-LF pairs, while \scholar is a database of user questions about academic publications, with automatically generated SQL that was checked by asking the user if the output was correct. \citet{finegan2018improving} further repurposed a complementary dataset split for semantic parsing evaluation on several dominant datasets, including \geo and \scholar. In this paper, we use the proposed query-split, which is more challenging to the semantic parsing models. To efficiently evaluate the semantic parsing execution performance, we choose to use distilled test suits, purposed by \citet{}, where there are 182 NL-LF pairs in \geo and 315 NL-LF pairs in \scholar.
\paragraph{Datasets.} The \geo dataset contains 877 pairs of NL-LF pairs about U.S. geographical information. On the other hand, \scholar contains pairs of NL-SQL regarding information about academic publications. \citet{finegan2018improving} proposed a dataset split for evaluating the compositional generalization capability of semantic parsers on several datasets, including \geo and \scholar. The proposed split, referred to as the query-split, presents a more challenging scenario for semantic parsing models. This paper utilizes the query-split, where the two test sets in our experiments include 182 NL-LF pairs from \geo and 315 NL-LF pairs from \scholar, respectively, during the evaluation of the prompt-based semantic parser.

\paragraph{Hyperparameters.}
\label{sec:app:api:hparam}
We sample at most 200 tokens from \codex with temperature 0, with the following strings used as stop tokens to halt generation: ``-{}-'', ``\textbackslash n\textbackslash n'', ``;'', ``\#''.

\paragraph{Model Versioning.}
\label{sec:app:api:versioning}
The version of the \davinci model referred to in this paper is as of the midpoint of the year 2022.
\section{Adversarial Examples}
\label{app:more_examples}
Table~\ref{tab:full_examples} lists the examples generated by all perturbation strategies.
\input{sections/tables/tab-examples.tex}

%% file: sections/tables/tab-examples.tex
\begin{table*}[!ht]\small \setlength{\tabcolsep}{7pt}
\centering
% \resizebox{1.0\textwidth}{!}{
\begin{tabular}{p{2.0cm}p{9cm}}
\toprule 
Linguistic \quad Phenomenon & Samples (\st{Strikethrough} = Original Text, \textcolor{red}{\textbf{red}} = Adversarial Perturbation) \\
\midrule
 \multirow{2}{*}{\shortstack{Typo \\ (Word-level)}} & \multirow{2}{*}{\textbf{NL}: what can you \st{tell}\textcolor{red}{\textbf{te11}} me about \st{the}\textcolor{red}{\textbf{th e}}  population of missouri} \\\\
  \midrule
   \multirow{2}{*}{\shortstack{Random Deletion \\ (Word-level)}} & \multirow{2}{*}{\textbf{NL}: \st{what} can you tell me \st{about} the population of missouri} \\\\
  \midrule
  
   \multirow{2}{*}{\shortstack{Random Swap \\ (Word-level)}} & \multirow{2}{*}{\textbf{NL}: what can you tell me \st{about}\textcolor{red}{\textbf{missouri}} the population of \st{missouri}\textcolor{red}{\textbf{about}}} \\\\
  \midrule
  
   \multirow{3}{*}{\shortstack{Context-aware\\Substitution\\ (Word-level)}} & \multirow{3}{*}{\textbf{NL}: what \st{can}\textcolor{red}{\textbf{will}} you tell me about \st{the}\textcolor{red}{\textbf{a}} population of missouri} \\\\\\
  \midrule
  
  \multirow{3}{*}{\shortstack{Context-aware\\ Insertion \\ (Word-level)}} & \multirow{3}{*}{\textbf{NL}: what \textcolor{red}{\textbf{what}} can you tell me about the \textcolor{red}{\textbf{exact}} population of missouri} \\\\\\
  \midrule

 \multirow{2}{*}{\shortstack{Rewriting \\ (Sent.-level)}} & \multirow{1}{*}{\textbf{NL}: \st{what can you tell me about the population of missouri}}  \\ & \textcolor{red}{\textbf{What information can you provide on Missouri's population?}}\\
 \midrule
 
  \multirow{2}{*}{\shortstack{Distraction \\ (Sent.-level)}} & \multirow{1}{*}{\textbf{NL}: what can you tell me about the population of missouri}  \\ & \textcolor{red}{\textbf{who is who; what is what; when is when; which is which; where is where}}\\
\bottomrule

\end{tabular}
% }
\caption{\textbf{Examples from Robustness Evaluation Set}. The adversarial utterances in this Table are generated by applying various perturbation strategies to a single utterance ``\textit{what can you tell me about the population of missouri}'' sampled from the \geo dataset. All of the generated utterances can successfully alter Codex's output.}
\label{tab:full_examples}
\end{table*}

%% file: custom.bib
@inproceedings{shin2021constrained,
  title={Constrained Language Models Yield Few-Shot Semantic Parsers},
  author={Shin, Richard and Lin, Christopher and Thomson, Sam and Chen Jr, Charles and Roy, Subhro and Platanios, Emmanouil Antonios and Pauls, Adam and Klein, Dan and Eisner, Jason and Van Durme, Benjamin},
  booktitle={Proceedings of the 2021 Conference on Empirical Methods in Natural Language Processing},
  pages={7699--7715},
  year={2021}
}

@article{roy2022benchclamp,
  title={BenchCLAMP: A Benchmark for Evaluating Language Models on Semantic Parsing},
  author={Roy, Subhro and Thomson, Sam and Chen, Tongfei and Shin, Richard and Pauls, Adam and Eisner, Jason and Van Durme, Benjamin},
  journal={arXiv preprint arXiv:2206.10668},
  year={2022}
}

@article{drozdov2022compositional,
  title={Compositional Semantic Parsing with Large Language Models},
  author={Drozdov, Andrew and Sch{\"a}rli, Nathanael and Akyu{\"u}rek, Ekin and Scales, Nathan and Song, Xinying and Chen, Xinyun and Bousquet, Olivier and Zhou, Denny},
  journal={arXiv preprint arXiv:2209.15003},
  year={2022}
}

@article{rajkumar2022evaluating,
  title={Evaluating the Text-to-SQL Capabilities of Large Language Models},
  author={Rajkumar, Nitarshan and Li, Raymond and Bahdanau, Dzmitry},
  journal={arXiv preprint arXiv:2204.00498},
  year={2022}
}

@inproceedings{wang2021adversarial,
  title={Adversarial GLUE: A Multi-Task Benchmark for Robustness Evaluation of Language Models},
  author={Wang, Boxin and Xu, Chejian and Wang, Shuohang and Gan, Zhe and Cheng, Yu and Gao, Jianfeng and Awadallah, Ahmed Hassan and Li, Bo},
  booktitle={Thirty-fifth Conference on Neural Information Processing Systems Datasets and Benchmarks Track (Round 2)},
  year={2021}
}

@inproceedings{huang2021robustness,
  title={On Robustness of Neural Semantic Parsers},
  author={Huang, Shuo and Li, Zhuang and Qu, Lizhen and Pan, Lei},
  booktitle={Proceedings of the 16th Conference of the European Chapter of the Association for Computational Linguistics: Main Volume},
  pages={3333--3342},
  year={2021}
}

@article{suzgun2022prompt,
  title={Prompt-and-rerank: A method for zero-shot and few-shot arbitrary textual style transfer with small language models},
  author={Suzgun, Mirac and Melas-Kyriazi, Luke and Jurafsky, Dan},
  journal={arXiv preprint arXiv:2205.11503},
  year={2022}
}

@book{templin1957certain,
  title={Certain language skills in children: Their development and interrelationships},
  author={Templin, Mildred C},
  volume={10},
  year={1957},
  publisher={JSTOR}
}

@book{wu2012advances,
  title={Advances in K-means clustering: a data mining thinking},
  author={Wu, Junjie},
  year={2012},
  publisher={Springer Science \& Business Media}
}

@misc{yule1944statistical,
  title={The statistical study of literary vocabulary. Cambridge, Cambridge [Eng.]},
  author={Yule, GU},
  year={1944},
  publisher={University Press. Journal of the Royal Statistical Society}
}

@phdthesis{mccarthy2005assessment,
  title={An assessment of the range and usefulness of lexical diversity measures and the potential of the measure of textual, lexical diversity (MTLD)},
  author={McCarthy, Philip M},
  year={2005},
  school={The University of Memphis}
}

@article{liu2021pre,
  title={Pre-train, prompt, and predict: A systematic survey of prompting methods in natural language processing},
  author={Liu, Pengfei and Yuan, Weizhe and Fu, Jinlan and Jiang, Zhengbao and Hayashi, Hiroaki and Neubig, Graham},
  journal={arXiv preprint arXiv:2107.13586},
  year={2021}
}

@inproceedings{
kojima2022large,
title={Large Language Models are Zero-Shot Reasoners},
author={Takeshi Kojima and Shixiang Shane Gu and Machel Reid and Yutaka Matsuo and Yusuke Iwasawa},
booktitle={ICML 2022 Workshop on Knowledge Retrieval and Language Models},
year={2022},
url={https://openreview.net/forum?id=6p3AuaHAFiN}
}

@article{marasovic2021few,
  title={Few-shot self-rationalization with natural language prompts},
  author={Marasovi{\'c}, Ana and Beltagy, Iz and Downey, Doug and Peters, Matthew E},
  journal={arXiv preprint arXiv:2111.08284},
  year={2021}
}

@inproceedings{ramesh2021zero,
  title={Zero-shot text-to-image generation},
  author={Ramesh, Aditya and Pavlov, Mikhail and Goh, Gabriel and Gray, Scott and Voss, Chelsea and Radford, Alec and Chen, Mark and Sutskever, Ilya},
  booktitle={International Conference on Machine Learning},
  pages={8821--8831},
  year={2021},
  organization={PMLR}
}

@inproceedings{yang2022empirical,
  title={An empirical study of gpt-3 for few-shot knowledge-based vqa},
  author={Yang, Zhengyuan and Gan, Zhe and Wang, Jianfeng and Hu, Xiaowei and Lu, Yumao and Liu, Zicheng and Wang, Lijuan},
  booktitle={Proceedings of the AAAI Conference on Artificial Intelligence},
  volume={36},
  pages={3081--3089},
  year={2022}
}

@inproceedings{yi2021improved,
  title={Improved ood generalization via adversarial training and pretraing},
  author={Yi, Mingyang and Hou, Lu and Sun, Jiacheng and Shang, Lifeng and Jiang, Xin and Liu, Qun and Ma, Zhiming},
  booktitle={International Conference on Machine Learning},
  pages={11987--11997},
  year={2021},
  organization={PMLR}
}

@inproceedings{bartolo2021improving,
  title={Improving Question Answering Model Robustness with Synthetic Adversarial Data Generation},
  author={Bartolo, Max and Thrush, Tristan and Jia, Robin and Riedel, Sebastian and Stenetorp, Pontus and Kiela, Douwe},
  booktitle={Proceedings of the 2021 Conference on Empirical Methods in Natural Language Processing},
  pages={8830--8848},
  year={2021}
}

@inproceedings{min2021curious,
  title={The curious case of adversarially robust models: More data can help, double descend, or hurt generalization},
  author={Min, Yifei and Chen, Lin and Karbasi, Amin},
  booktitle={Uncertainty in Artificial Intelligence},
  pages={129--139},
  year={2021},
  organization={PMLR}
}

@article{raghunathan2019adversarial,
  title={Adversarial training can hurt generalization},
  author={Raghunathan, Aditi and Xie, Sang Michael and Yang, Fanny and Duchi, John C and Liang, Percy},
  journal={arXiv preprint arXiv:1906.06032},
  year={2019}
}

@inproceedings{sanghi2022clip,
  title={Clip-forge: Towards zero-shot text-to-shape generation},
  author={Sanghi, Aditya and Chu, Hang and Lambourne, Joseph G and Wang, Ye and Cheng, Chin-Yi and Fumero, Marco and Malekshan, Kamal Rahimi},
  booktitle={Proceedings of the IEEE/CVF Conference on Computer Vision and Pattern Recognition},
  pages={18603--18613},
  year={2022}
}

@inproceedings{guo2021gradient,
  title={Gradient-based Adversarial Attacks against Text Transformers},
  author={Guo, Chuan and Sablayrolles, Alexandre and J{\'e}gou, Herv{\'e} and Kiela, Douwe},
  booktitle={Proceedings of the 2021 Conference on Empirical Methods in Natural Language Processing},
  pages={5747--5757},
  year={2021}
}

@inproceedings{ross2022tailor,
  title={Tailor: Generating and Perturbing Text with Semantic Controls},
  author={Ross, Alexis and Wu, Tongshuang and Peng, Hao and Peters, Matthew E and Gardner, Matt},
  booktitle={Proceedings of the 60th Annual Meeting of the Association for Computational Linguistics (Volume 1: Long Papers)},
  pages={3194--3213},
  year={2022}
}

@article{hosseini2017deceiving,
  title={Deceiving google's perspective api built for detecting toxic comments},
  author={Hosseini, Hossein and Kannan, Sreeram and Zhang, Baosen and Poovendran, Radha},
  journal={arXiv preprint arXiv:1702.08138},
  year={2017}
}

@inproceedings{ebrahimi2018hotflip,
  title={HotFlip: White-Box Adversarial Examples for Text Classification},
  author={Ebrahimi, Javid and Rao, Anyi and Lowd, Daniel and Dou, Dejing},
  booktitle={Proceedings of the 56th Annual Meeting of the Association for Computational Linguistics (Volume 2: Short Papers)},
  pages={31--36},
  year={2018}
}

@inproceedings{belinkov2018synthetic,
  title={Synthetic and Natural Noise Both Break Neural Machine Translation},
  author={Belinkov, Yonatan and Bisk, Yonatan},
  booktitle={International Conference on Learning Representations},
  year={2018}
}

@inproceedings{gao2018black,
  title={Black-box generation of adversarial text sequences to evade deep learning classifiers},
  author={Gao, Ji and Lanchantin, Jack and Soffa, Mary Lou and Qi, Yanjun},
  booktitle={2018 IEEE Security and Privacy Workshops (SPW)},
  pages={50--56},
  year={2018},
  organization={IEEE}
}

@inproceedings{eger2019text,
  title={Text Processing Like Humans Do: Visually Attacking and Shielding NLP Systems},
  author={Eger, Steffen and {\c{S}}ahin, G{\"o}zde G{\"u}l and R{\"u}ckl{\'e}, Andreas and Lee, Ji-Ung and Schulz, Claudia and Mesgar, Mohsen and Swarnkar, Krishnkant and Simpson, Edwin and Gurevych, Iryna},
  booktitle={Proceedings of the 2019 Conference of the North American Chapter of the Association for Computational Linguistics: Human Language Technologies, Volume 1 (Long and Short Papers)},
  pages={1634--1647},
  year={2019}
}

@inproceedings{boucher2022bad,
  title={Bad characters: Imperceptible nlp attacks},
  author={Boucher, Nicholas and Shumailov, Ilia and Anderson, Ross and Papernot, Nicolas},
  booktitle={2022 IEEE Symposium on Security and Privacy (SP)},
  pages={1987--2004},
  year={2022},
  organization={IEEE}
}

@inproceedings{iyyer2018adversarial,
  title={Adversarial Example Generation with Syntactically Controlled Paraphrase Networks},
  author={Iyyer, Mohit and Wieting, John and Gimpel, Kevin and Zettlemoyer, Luke},
  booktitle={Proceedings of the 2018 Conference of the North American Chapter of the Association for Computational Linguistics: Human Language Technologies, Volume 1 (Long Papers)},
  pages={1875--1885},
  year={2018}
}

@article{miyato2016adversarial,
  title={Adversarial training methods for semi-supervised text classification},
  author={Miyato, Takeru and Dai, Andrew M and Goodfellow, Ian},
  journal={arXiv preprint arXiv:1605.07725},
  year={2016}
}

@inproceedings{ribeiro2018semantically,
  title={Semantically equivalent adversarial rules for debugging NLP models},
  author={Ribeiro, Marco Tulio and Singh, Sameer and Guestrin, Carlos},
  booktitle={Annual Meeting of the Association for Computational Linguistics (ACL)},
  year={2018}
}

@inproceedings{zhao2018generating,
  title={Generating Natural Adversarial Examples},
  author={Zhao, Zhengli and Dua, Dheeru and Singh, Sameer},
  booktitle={International Conference on Learning Representations},
  year={2018}
}

@inproceedings{alzantot2018generating,
  title={Generating Natural Language Adversarial Examples},
  author={Alzantot, Moustafa and Sharma, Yash and Elgohary, Ahmed and Ho, Bo-Jhang and Srivastava, Mani and Chang, Kai-Wei},
  booktitle={Proceedings of the 2018 Conference on Empirical Methods in Natural Language Processing},
  pages={2890--2896},
  year={2018}
}

@inproceedings{liang2018deep,
  title={Deep Text Classification Can be Fooled},
  author={Liang, Bin and Li, Hongcheng and Su, Miaoqiang and Bian, Pan and Li, Xirong and Shi, Wenchang},
  booktitle={IJCAI},
  year={2018}
}

@inproceedings{zhang2019theoretically,
  title={Theoretically principled trade-off between robustness and accuracy},
  author={Zhang, Hongyang and Yu, Yaodong and Jiao, Jiantao and Xing, Eric and El Ghaoui, Laurent and Jordan, Michael},
  booktitle={International conference on machine learning},
  pages={7472--7482},
  year={2019},
  organization={PMLR}
}

@inproceedings{szegedy2013intriguing,
  author    = {Christian Szegedy and
               Wojciech Zaremba and
               Ilya Sutskever and
               Joan Bruna and
               Dumitru Erhan and
               Ian J. Goodfellow and
               Rob Fergus},
  editor    = {Yoshua Bengio and
               Yann LeCun},
  title     = {Intriguing properties of neural networks},
  booktitle = {2nd International Conference on Learning Representations, {ICLR} 2014,
               Banff, AB, Canada, April 14-16, 2014, Conference Track Proceedings},
  year      = {2014},
  url       = {http://arxiv.org/abs/1312.6199},
  bibsource = {dblp computer science bibliography, https://dblp.org}
}

@inproceedings{shin-van-durme-2022-shot,
    title = "Few-Shot Semantic Parsing with Language Models Trained on Code",
    author = "Shin, Richard  and
      Van Durme, Benjamin",
    booktitle = "Proceedings of the 2022 Conference of the North American Chapter of the Association for Computational Linguistics: Human Language Technologies",
    month = jul,
    year = "2022",
    address = "Seattle, United States",
    publisher = "Association for Computational Linguistics",
    url = "https://aclanthology.org/2022.naacl-main.396",
    doi = "10.18653/v1/2022.naacl-main.396",
    pages = "5417--5425"
}

@misc{BenchCLAMP,
  doi = {10.48550/ARXIV.2206.10668},
  
  url = {https://arxiv.org/abs/2206.10668},
  
  author = {Roy, Subhro and Thomson, Sam and Chen, Tongfei and Shin, Richard and Pauls, Adam and Eisner, Jason and Van Durme, Benjamin},
  
  title = {BenchCLAMP: A Benchmark for Evaluating Language Models on Semantic Parsing},
  
  publisher = {arXiv},
  
  year = {2022},
  
  copyright = {Creative Commons Attribution 4.0 International}
}

@inproceedings{allen2022feature,
  title={Feature purification: How adversarial training performs robust deep learning},
  author={Allen-Zhu, Zeyuan and Li, Yuanzhi},
  booktitle={2021 IEEE 62nd Annual Symposium on Foundations of Computer Science (FOCS)},
  pages={977--988},
  year={2022},
  organization={IEEE}
}

@article{li2018textbugger,
  title={Textbugger: Generating adversarial text against real-world applications},
  author={Li, Jinfeng and Ji, Shouling and Du, Tianyu and Li, Bo and Wang, Ting},
  journal={arXiv preprint arXiv:1812.05271},
  year={2018}
}

@article{naik2018stress,
  title={Stress test evaluation for natural language inference},
  author={Naik, Aakanksha and Ravichander, Abhilasha and Sadeh, Norman and Rose, Carolyn and Neubig, Graham},
  journal={arXiv preprint arXiv:1806.00692},
  year={2018}
}

@article{fitria2021quillbot,
  title={QuillBot as an online tool: Students’ alternative in paraphrasing and rewriting of English writing},
  author={Fitria, Tira Nur},
  journal={Englisia: Journal of Language, Education, and Humanities},
  volume={9},
  number={1},
  pages={183--196},
  year={2021}
}

@inproceedings{shiri2022paraphrasing,
  title={Paraphrasing Techniques for Maritime QA system},
  author={Shiri, Fatemeh and Zhuo, Terry Yue and Li, Zhuang and Pan, Shirui and Wang, Weiqing and Haffari, Reza and Li, Yuan-Fang and Nguyen, Van},
  booktitle={2022 25th International Conference on Information Fusion (FUSION)},
  pages={1--8},
  year={2022},
  organization={IEEE}
}

@article{liu2019roberta,
  title={Roberta: A robustly optimized bert pretraining approach},
  author={Liu, Yinhan and Ott, Myle and Goyal, Naman and Du, Jingfei and Joshi, Mandar and Chen, Danqi and Levy, Omer and Lewis, Mike and Zettlemoyer, Luke and Stoyanov, Veselin},
  journal={arXiv preprint arXiv:1907.11692},
  year={2019}
}

@inproceedings{li2021total,
  title={Total Recall: a Customized Continual Learning Method for Neural Semantic Parsers},
  author={Li, Zhuang and Qu, Lizhen and Haffari, Gholamreza},
  booktitle={Proceedings of the 2021 Conference on Empirical Methods in Natural Language Processing},
  pages={3816--3831},
  year={2021}
}

@inproceedings{reimers2019sentence,
  title={Sentence-BERT: Sentence Embeddings using Siamese BERT-Networks},
  author={Reimers, Nils and Gurevych, Iryna},
  booktitle={Proceedings of the 2019 Conference on Empirical Methods in Natural Language Processing and the 9th International Joint Conference on Natural Language Processing (EMNLP-IJCNLP)},
  pages={3982--3992},
  year={2019}
}

@inproceedings{papineni2002bleu,
  title={Bleu: a method for automatic evaluation of machine translation},
  author={Papineni, Kishore and Roukos, Salim and Ward, Todd and Zhu, Wei-Jing},
  booktitle={Proceedings of the 40th annual meeting of the Association for Computational Linguistics},
  pages={311--318},
  year={2002}
}

@inproceedings{poon2009unsupervised,
  title={Unsupervised semantic parsing},
  author={Poon, Hoifung and Domingos, Pedro},
  booktitle={Proceedings of the 2009 conference on empirical methods in natural language processing},
  pages={1--10},
  year={2009}
}

@article{chen2021evaluating,
  title={Evaluating large language models trained on code},
  author={Chen, Mark and Tworek, Jerry and Jun, Heewoo and Yuan, Qiming and Pinto, Henrique Ponde de Oliveira and Kaplan, Jared and Edwards, Harri and Burda, Yuri and Joseph, Nicholas and Brockman, Greg and others},
  journal={arXiv preprint arXiv:2107.03374},
  year={2021}
}

@inproceedings{devlin2019bert,
  title={BERT: Pre-training of Deep Bidirectional Transformers for Language Understanding},
  author={Devlin, Jacob and Chang, Ming-Wei and Lee, Kenton and Toutanova, Kristina},
  booktitle={Proceedings of the 2019 Conference of the North American Chapter of the Association for Computational Linguistics: Human Language Technologies, Volume 1 (Long and Short Papers)},
  pages={4171--4186},
  year={2019}
}

@inproceedings{kamath2018survey,
  title={A Survey on Semantic Parsing},
  author={Kamath, Aishwarya and Das, Rajarshi},
  booktitle={Automated Knowledge Base Construction (AKBC)},
  year={2018}
}

@article{bapna2017towards,
  title={Towards zero-shot frame semantic parsing for domain scaling},
  author={Bapna, Ankur and Tur, Gokhan and Hakkani-Tur, Dilek and Heck, Larry},
  journal={arXiv preprint arXiv:1707.02363},
  year={2017}
}

@misc{gpt-j,
  author = {Wang, Ben and Komatsuzaki, Aran},
  title = {{GPT-J-6B: A 6 Billion Parameter Autoregressive Language Model}},
  howpublished = {\url{https://github.com/kingoflolz/mesh-transformer-jax}},
  year = 2021,
  month = May
}

@inproceedings{zelle1996learning,
  title={Learning to parse database queries using inductive logic programming},
  author={Zelle, John M and Mooney, Raymond J},
  booktitle={Proceedings of the national conference on artificial intelligence},
  pages={1050--1055},
  year={1996}
}

@inproceedings{iyer2017learning,
  title={Learning a Neural Semantic Parser from User Feedback},
  author={Iyer, Srinivasan and Konstas, Ioannis and Cheung, Alvin and Krishnamurthy, Jayant and Zettlemoyer, Luke},
  booktitle={Proceedings of the 55th Annual Meeting of the Association for Computational Linguistics (Volume 1: Long Papers)},
  pages={963--973},
  year={2017}
}

@inproceedings{pi2022towards,
  title={Towards Robustness of Text-to-SQL Models Against Natural and Realistic Adversarial Table Perturbation},
  author={Pi, Xinyu and Wang, Bing and Gao, Yan and Guo, Jiaqi and Li, Zhoujun and Lou, Jian-Guang},
  booktitle={Proceedings of the 60th Annual Meeting of the Association for Computational Linguistics (Volume 1: Long Papers)},
  pages={2007--2022},
  year={2022}
}

@article{shafahi2019adversarial,
  title={Adversarial training for free!},
  author={Shafahi, Ali and Najibi, Mahyar and Ghiasi, Mohammad Amin and Xu, Zheng and Dickerson, John and Studer, Christoph and Davis, Larry S and Taylor, Gavin and Goldstein, Tom},
  journal={Advances in Neural Information Processing Systems},
  volume={32},
  year={2019}
}

@article{ganin2016domain,
  title={Domain-adversarial training of neural networks},
  author={Ganin, Yaroslav and Ustinova, Evgeniya and Ajakan, Hana and Germain, Pascal and Larochelle, Hugo and Laviolette, Fran{\c{c}}ois and Marchand, Mario and Lempitsky, Victor},
  journal={The journal of machine learning research},
  volume={17},
  number={1},
  pages={2096--2030},
  year={2016},
  publisher={JMLR. org}
}

@inproceedings{shafahi2020universal,
  title={Universal adversarial training},
  author={Shafahi, Ali and Najibi, Mahyar and Xu, Zheng and Dickerson, John and Davis, Larry S and Goldstein, Tom},
  booktitle={Proceedings of the AAAI Conference on Artificial Intelligence},
  volume={34},
  pages={5636--5643},
  year={2020}
}

@article{tramer2019adversarial,
  title={Adversarial training and robustness for multiple perturbations},
  author={Tramer, Florian and Boneh, Dan},
  journal={Advances in Neural Information Processing Systems},
  volume={32},
  year={2019}
}

@inproceedings{duong2018active,
  title={Active learning for deep semantic parsing},
  author={Duong, Long and Afshar, Hadi and Estival, Dominique and Pink, Glen and Cohen, Philip R and Johnson, Mark},
  booktitle={Proceedings of the 56th Annual Meeting of the Association for Computational Linguistics (Volume 2: Short Papers)},
  pages={43--48},
  year={2018}
}

@inproceedings{sen-yilmaz-2020-uncertainty,
    title = "Uncertainty and Traffic-Aware Active Learning for Semantic Parsing",
    author = "Sen, Priyanka  and
      Yilmaz, Emine",
    booktitle = "Proceedings of the First Workshop on Interactive and Executable Semantic Parsing",
    month = nov,
    year = "2020",
    address = "Online",
    publisher = "Association for Computational Linguistics",
    url = "https://aclanthology.org/2020.intexsempar-1.2",
    doi = "10.18653/v1/2020.intexsempar-1.2",
    pages = "12--17",
}

@inproceedings{pang2019improving,
  title={Improving adversarial robustness via promoting ensemble diversity},
  author={Pang, Tianyu and Xu, Kun and Du, Chao and Chen, Ning and Zhu, Jun},
  booktitle={International Conference on Machine Learning},
  pages={4970--4979},
  year={2019},
  organization={PMLR}
}

@inproceedings{finegan2018improving,
  title={Improving Text-to-SQL Evaluation Methodology},
  author={Finegan-Dollak, Catherine and Kummerfeld, Jonathan K and Zhang, Li and Ramanathan, Karthik and Sadasivam, Sesh and Zhang, Rui and Radev, Dragomir},
  booktitle={Proceedings of the 56th Annual Meeting of the Association for Computational Linguistics (Volume 1: Long Papers)},
  pages={351--360},
  year={2018}
}

@article{wagner1974string,
  title={The string-to-string correction problem},
  author={Wagner, Robert A and Fischer, Michael J},
  journal={Journal of the ACM (JACM)},
  volume={21},
  number={1},
  pages={168--173},
  year={1974},
  publisher={ACM New York, NY, USA}
}

@book{anand2011mining,
  title={Mining of massive datasets},
  author={Anand, Rajaraman and Jeffrey David, Ullman},
  year={2011},
  publisher={Cambridge University Press}
}

@article{zhuo2023exploring,
  title={Exploring ai ethics of chatgpt: A diagnostic analysis},
  author={Zhuo, Terry Yue and Huang, Yujin and Chen, Chunyang and Xing, Zhenchang},
  journal={arXiv preprint arXiv:2301.12867},
  year={2023}
}
